# An Anthropic Argument against the Future Existence of Superintelligent Artificial Intelligence

Toby Pereira

8th May 2017

## Abstract

This paper uses anthropic reasoning to argue for a reduced likelihood that superintelligent AI will come into existence in the future. To make this argument, a new principle is introduced: the Super-Strong Self-Sampling Assumption (SSSSA), building on the Self-Sampling Assumption (SSA) and the Strong Self-Sampling Assumption (SSSA). SSA uses as its sample the relevant observers, whereas SSSA goes further by using observer-moments. SSSSA goes further still and weights each sample proportionally, according to the size of a mind in cognitive terms. SSSSA is required for human observer-samples to be typical, given by how much non-human animals outnumber humans. Given SSSSA, the assumption that humans experience typical observer-samples relies on a future where superintelligent AI does not dominate, which in turn reduces the likelihood of it being created at all.

## 1. Introduction to Anthropic Reasoning

The fact that we exist on Earth, a life-permitting planet, might superficially seem like a stroke of luck. However, it couldn't really be any other way. All conscious observers must exist in a place compatible with their existence. So if there is life at all in the universe, the fact that it will be experiencing such a planet is inevitable. Similarly, we can only exist at a time in the universe's and the Earth's history when conditions are right for life. So if life can only exist for a small slice of time, either on Earth or in the universe as a whole, we should not be surprised or feel lucky that that time is now.

But this is not just about absolutes – places and times where life can and cannot exist. Places and times can exist on a scale from very life-friendly to very life-hostile. As a hypothetical (and unrealistic) example, we could find out that there are several galaxies where life exists. But among these, some could be more conducive to life than others and have many more planets with life on them. To keep things simple, imagine that there are two types of galaxy, and that these two types are equally common. One type is relatively life-friendly and averages 1000 planets with life per galaxy. The other is more life-hostile and averages one planet with life per galaxy. For this example, we will also assume that the make-up of the planets with life in each type of galaxy is roughly similar, with the same probability of different types of life evolving and the same average number of living organisms etc.

Assuming that we didn't already know which type of galaxy the Milky Way was, we would reason that there is approximately a 99.9% chance that it is the more life-friendly type. If most conscious observers are in a certain type of galaxy, then you, as a conscious observer, should reason that you are more likely to be in this type of galaxy, given no other information. All other things being equal, you should expect yourself to be in the more typical situation.

This goes further than expecting your environment to be typical. It is also about expecting yourself to be a typical observer. For example, imagine that there are two variants of a particular gene, which cause people to see colours very slightly differently, and because the differences are so small, no-one knows which variant they have unless they go through a rigorous sight test, or indeed a genetic test.

If 95% of people had one gene variant and 5% the other, then with no other information about your own situation, you should reason that you are more likely to have the more common variant. And, as advertised, this is part of you, rather than your environment.

The flip side of this is that if you didn't know which was the more common variant, but you did know which variant you had, you would reason that yours was probably the more common variant (with 95% probability). This would then allow you to make predictions about which variants other people had (you would expect most people to have the same variant as you). This is an important point because it shows that you can make predications about the rest of the world from your own case rather than just the other way round. And this is the basis of anthropic reasoning.

## 2. SSA, SSSA, SSSSA and Boltzmann Brains

This brings us to the Self-Sampling Assumption (SSA), defined by Nick Bostrom as follows:

> (SSA) One should reason as if one were a random sample from the set of all observers in one's reference class. (2002, p. 57)

The reference class is the class of entities that you should consider yourself to be a sample from. When considering the probability of having a particular gene variant in the example above, the statistics relate to humans, and you are a human yourself, so a sensible reference class to use would be that of all humans. But you might also happen to know that eye colour affects the probability, so you could narrow down the reference class further and include only people with the same eye colour as you.

When evaluating your position as a conscious observer more generally, a more general reference class would be required. For example, if there are many other races of advanced intelligent life in the galaxy (life that has developed speech and writing, say), then we could make predictions about them on the basis that we would expect human life to be fairly typical in most respects among this intelligent life.

Depending on what question you ask, the reference class you use could potentially stretch across the whole universe, backwards and forwards in time, and even into any other universes that might exist. And given that we would expect a random sample to be fairly typical in most respects, we can use our own case to make predictions about what or who else is out there, across the whole universe and beyond. For example, if there are other universes out there, causally unconnected to our own, and some of these have intelligent life, we would expect the intelligent life in this universe to be fairly typical, given no information to the contrary. This would enable us to begin to make predictions about the life in these other universes.

Anthropic reasoning of this sort is sometimes used to evaluate theories in physics. If we find ourselves faced with a theory of reality that leads to us being very atypical conscious observers in the universe, it is arguably grounds to be suspicious of that theory. For example, something that often worries physicists is the idea of Boltzmann Brains, named after the physicist and philosopher Ludwig Boltzmann. According to the second law of thermodynamics, entropy (roughly a measure of disorder) only increases, or at least doesn't decrease. But this is a statistical law, and there can be local fluctuations where entropy can decrease by random chance, causing a more ordered state. Most of these fluctuations will be very small and insignificant, and the newly found order will quickly return to disorder.

But wait long enough and it is statistically probable that there will be a large local decrease in entropy, and a complex object will randomly fluctuate into existence. And if you wait a really long time, a fully-formed human brain complete with its own thoughts and memories will fluctuate into existence. Such brains are known as Boltzmann Brains.

If a theory of physics leads to the conclusion that Boltzmann Brains significantly outnumber human-like brains that have evolved normally on a planet such as Earth, then a consequence of this physical theory is that human beings would be highly atypical observers. But by SSA, we should reason that we probably are fairly typical observers, so this seems to suggest that Boltzmann Brains do not in fact significantly outnumber normal human-like brains. Therefore, this would give us grounds to doubt the theory of physics.

However, this should not lead us to start worrying that we are in fact Boltzmann Brains but without knowing it. The vast majority Boltzmann Brains would not be experiencing anything like a coherent universe (given that their make-up is effectively random), whereas we are. If we were Boltzmann Brains, we would be highly atypical ones, which should give us strong grounds to doubt that we are.

It might seem that these Boltzmann Brains could never pose a serious problem. They would clearly be very few and far between, whereas there are billions of humans on planet Earth. There is also the possibility of human-like intelligent beings existing on many other planets throughout the universe. So a few of these Boltzmann Brains fluctuating in and out of existence should have no bearing on whether we, as normal humans on planet Earth, are typical observers.

However, according to some theories, entropy will continually increase, until eventually the universe reaches thermal equilibrium, or "heat death". At this point, all the matter in the universe will be uniformly spread out, with no galaxies, planets, or indeed life. Except, that is, for random fluctuations. And if the universe simply exists eternally in this state of heat death, then however rare these Boltzmann Brain fluctuations are, eventually they will come to outnumber "normal" brains, and infinitely so.

This is a reason why some physicists reason that the universe cannot be like this, and that there must be some other outcome, such as the universe ending in a big crunch after a finite time, or a static end-point where no fluctuations can come about. See Carroll (2017) for a recent discussion of Boltzmann Brains in physics.

A possible defence against the threat of Boltzmann Brains is that they could not survive in the vacuum of outer space for more than a few seconds. This means that even if they outnumber human brains when looking at absolute numbers, they may not do so when it comes to "observer-moments". This brings us to Bostrom's modification of SSA, the Strong Self-Sampling Assumption (SSSA):

> (SSSA) One should reason as if one's present observer-moment were a random sample from the set of all observer-moments in its reference class. (2002, p. 162)

This is arguably a more sophisticated version of SSA. If a conscious mind exists only for a fraction of a second, then it is responsible for far less experience than one that exists for 80 years, and it makes sense that this should be taken into account when determining how to define an observer-sample.

But of course a Boltzmann Brain does not need to survive for more than the tiniest fraction of a second if there are enough of them, as would be the case if the heat-death state of the universe is eternal. Their observer-moments would still infinitely outnumber those of normal brains. But this is an aside, as this purported defence against the threat of Boltzmann Brains was really just a way to introduce SSSA!

So far, so good. However, it would appear that we are actually not typical conscious observers, and that our observer-moments are not typical observer-moments by any stretch of the imagination, and we don't need to speculate into the depths of space and time to see this.

We are just one of millions of species of animal on planet Earth, with myriad individual animals among these. It seems reasonable to suggest that the individuals of many of these species have some degree of consciousness themselves. Humans would, therefore, only comprise a tiny proportion of the total number of conscious beings on planet Earth, even if we limited ourselves to vertebrates or even just mammals. And being far more cognitively advanced than the animals of other species, we are not typical conscious observers, but incredibly privileged ones. So does this present a problem for our anthropic reasoning? How can we expect the logic to hold that we should be typical conscious observers throughout time and space, when we are not typical conscious observers even on our own planet right now?

There is a possible solution to this. While we may comprise only a tiny fraction of conscious beings on planet Earth, our superior intelligence and cognitive power that seemingly makes us so atypical means that we arguably have a greater amount of conscious experience than other animals. It could be said that our minds take up a larger area of "consciousness-space", suitably defined, than the minds of other animals.

So instead of picking a conscious observer-moment at random, if we took a random point in "consciousness-space-time" and used that as our observer-sample, we would be far more likely to find ourselves in a human mind.

Bostrom seems to have hit upon the same idea, but without fully following it through. He points out that if a being has a sped-up brain (like a computer running at a faster clock speed), then it will experience more within a given time than another being. Bostrom suggests that such a being would have more observer-moments per second. However, arguably its main implication is relegated to a footnote.

> One can ponder whether one should not also assign a higher sampling density to certain types of observer-moments, for example those that have a greater degree of clarity, intensity, or focus. (2002, p. 165)

Indeed, a mind with a faster clock speed is just one way of having more conscious experience per unit time. And this brings us to a further extension of SSA and SSSA:

Super-Strong Self-Sampling Assumption (SSSSA) – One should reason as if one's present observer-moment were a random sample from the set of all observer-moments in its reference class, with the probabilities weighted proportionally according to observer-moments' size in consciousness-space-time.

This adds another layer of sophistication onto SSSA by taking into account all of the conscious experience in the reference class together, rather than by separating it out into discrete units of vastly different sizes (such as a mouse mind and a human mind) and giving all these units the same probability weighting. It makes sense to say that human consciousness is more typical than mouse consciousness if there is more of it in total, regardless of how many discrete units there are. As an analogy, water is more typical on the surface of the Earth than land, even though there are more countries than oceans.

Using SSSSA also means that there isn't such a discontinuity at the transition between having no consciousness and having a tiny bit of consciousness. Using SSA or SSSA would mean that as soon as a being develops any consciousness, its observer-samples abruptly go from having zero probability weighting to having a full weighting, even if the amount of consciousness is negligible. Using SSSSA, there is a smooth transition where probability weighting is proportional to amount of consciousness.

In this situation, where we are wondering why we are ourselves as opposed to any other type of conscious being, it makes sense to consider the widest possible reference class: that of all conscious observers across all space and time.

Of course, it is still an open question as to whether human consciousness does take up enough of consciousness-space-time to make up for the lack of individual humans when compared to other conscious animals, but it could arguably be tested if we had a workable measure of the size of a being's consciousness.

The Integrated Information Theory of consciousness has a measure of consciousness, called Φ (the Greek letter 'phi'). See e.g. Tononi (2012); Oizumi, Albantakis & Tononi (2014). In his book *Phi: A Voyage from the Brain to the Soul*, Tononi (ibid.) introduces the idea of a qualiascope, which is a device that enables you to see objects in terms of their conscious properties rather than their physical properties. Viewed through a qualiascope, the size of an object is proportional to the amount of consciousness it has (its measure of Φ) rather than its physical size. This measure, or something along these lines, could be what we're looking for when determining a definition of size in consciousness-space.

SSSSA also allows us to predict that human-level intelligence evolves fairly frequently on planets where animal-like life evolves. As discussed, when considering whether we are typical conscious observers, we must consider all life in the universe. If human-level intelligence is very rare on planets where animal-like life has evolved, then picking a random point in consciousness-space-time is very unlikely to find a human-level intelligent mind, putting us in an improbably privileged position. This is why we would expect human-level intelligence to be relatively common.

## 3. Arguments against, and a Defence of, SSA, SSSA and SSSSA

SSA, SSSA and SSSSA seem to rely on the idea that our identity is somehow picked at random and that we could have been someone else. And one could rebut this by simply saying that something can only ever be itself, and wondering why it is this way is just meaningless. For example, you wouldn't look at a sheep and say that it could have been a cow. Similarly, it shouldn't matter how many of some other type of conscious being there are, because they don't have any relevant causal effect on you and your own identity.

A full discussion of the question "Why am I me?" and of personal identity generally could get very philosophical and is beyond the scope of this paper, but briefly, the claim is not that your identity was picked at random in this way, but reasoning *as if* you are a random observer-sample does give us a good methodology that allows us to assess probabilities in a reasonable manner.

I have already given examples where it seems correct to use these principles, such as the case with the gene variants that are possessed by 5% and 95% of the population respectively. The alternative is to say that you are who you are and that it makes no sense to put a probability on it. This seems to be a very unsatisfactory way to reason in this case.

Similarly, your confidence that you are not a Boltzmann Brain relies on anthropic reasoning of this sort. If we dismiss this reasoning, then it doesn't matter that a randomly formed brain in the vacuum of outer space is very unlikely to be experiencing such a coherent illusion of an ordered world and universe. It is simply a brute fact whether you are the mind of a Boltzmann Brain or the mind of a brain of a normal human being that has evolved on planet Earth, and it makes no sense to assign a probability to it. But clearly you live your life as if you are a normal human being, and probably do not consider the possibility that your reality will crumble away in a matter of moments, and you also probably think that this is rational as much as it is habitual.

Time-based examples that use SSA, SSSA or SSSSA are arguably even more counterintuitive. Even if one accepts that human-like minds should take up a reasonable proportion of consciousness-space at this particular time, to extend this to all time, particularly the future, could seem absurd, because the future hasn't happened yet and doesn't exist, so could have no bearing on what we should be experiencing now. There are various competing physical theories of time, some of which are in opposition to this claim, but such a discussion is beyond the scope of this paper. However, a defence of these time-based examples is possible without an appeal to physics. John Leslie gives the following example:

> Imagine an experiment planned as follows. At some point in time, three humans would each be given an emerald. Several centuries afterwards, when a completely different set of humans was alive, five thousand humans would each be given an emerald. Imagine next that you have yourself been given an emerald in the experiment. You have no knowledge, however, of whether your century is the earlier century in which just three people were to be in this situation, or in the later century in which five thousand were to be in it. Do you say to yourself that if yours were the earlier century then the five thousand people *wouldn't be alive yet*, and that therefore you'd have no chance of being among them? On this basis, do you conclude that you might just as well bet that you lived in the earlier century?
>
> Suppose you in fact betted that you lived there. If every emerald-getter in the experiment betted in this way, there would be five thousand losers and only three winners. The sensible bet, therefore, is that yours is instead the later century of the two. (1996, p. 20, italics in original)

Bostrom also defends the idea of making predictions about observers causally unconnected to you, based on your own experiences:

> To see why this "dependence on remote regions" is not a problem, it suffices to note that the probabilities our theory delivers are not physical chances but subjective credences. Those distant observers have zilch effect on the physical chances of events that take place on Earth. Rather, what holds is that under certain special circumstances, *your beliefs* about the distant observers could come to rationally affect *your beliefs* about about a nearby coin toss, say. (2002, p. 120, italics in original)

This paper will continue on the assumption that SSSSA is a sound principle.

## 4. Superintelligent Artificial Intelligence and the Argument against its Future Existence

By SSSSA, we would not expect to find ourselves in an improbably privileged position by being far more intelligent than the mind of an average point in consciousness-space-time, but the flip side of this is that we would also not expect ourselves to be in an improbably impoverished position, by being far less intelligent than the mind of an average point in consciousness-space-time. As well as calling into question the abundance of superintelligent alien life forms in the universe, this brings us neatly to superintelligent artificial intelligence (AI).

At some point in the future, we might develop a superintelligent AI that dwarfs our own intelligence. David Chalmers (2010) defines AI++ as AI that dwarfs the intelligence of the most intelligent human by at least as much as this human's intelligence dwarfs that of a mouse, and it seems sensible to use the same terminology here. A rigorously defined scale of intelligence would be required for this, but this should be achievable. From now on in this paper, AI++ and superintelligent AI can be considered interchangeable terms. If AI++ is conscious, we would expect the mind of an AI++ to take up a much greater area in consciousness-space than a human mind. It also seems likely that if one AI++ is created, then many will be created. And if many are created, then we might also expect them to dominate total consciousness-space-time.

But by SSSSA, we should expect our human minds to be the minds of fairly typical points in consciousness-space-time. This would mean that we should not expect AI++s to dominate consciousness-space-time. Given that we should not expect them to dominate, this also arguably reduces the likelihood of any AI++s being created at all. This is the central argument of the paper, and we can put it more formally:

Premise 1: If we create a single example of AI++, then many will be created, either by us or other AI++s.
Premise 2: If many AI++s are created, then they will dominate consciousness-space-time.
Premise 3: By SSSSA, AI++s probably do not dominate consciousness-space-time.

Conclusion: We will probably not create AI++.

It could be that instead of many individual AI++s being created, just one is created, and any further advances simply contribute to the further enhancement of this one AI++. While this means that premise 1 would not be correct as it stood, this example of a single super AI++ (AI+++, perhaps?) is still likely to dominate consciousness-space-time just as much as if there are many lesser AI++s. As it does not affect the conclusion, I will continue with the original premise.

Accepting this argument also has the knock-on effect of decreasing the likelihood that we will create even human-level AI. Creating human-level AI is a precursor to creating AI++, and assuming that AI++ is not created, we don't know where along the line that we will hit our limits. If we are told that a road is less than 100 miles long, then this also decreases the probability that it is at least 10 miles long, compared to before we had been given this information. The proportion of roads that are at least 10 miles long but less than 100 miles long is less than the proportion of roads that are at least 10 miles long without this upper limit. Similarly, the probability of creating human-level AI given that we are unlikely to create AI++ is less than the probability that we will create human-level AI without this extra information. Of course, this is relative and it gives us no absolute probability figure of us creating human-level AI in the future, and it could still be argued to be over 50%.

## 5. Attacks on Premise 1 – Simulations

If a single AI++ is created, then the likelihood of AI++s dominating consciousness-space-time does seem to be high, but it might not be inevitable.

Looking at premise 1, the idea that many AI++s will be created and that they will dominate the world, perhaps even wiping us out in the process, is just one possibility. It might be that relatively few are created and that they can co-exist with us. (Although, even if there are a relatively small number of individual AI++s created, one might still expect them to outlast the relatively fragile human race and indeed animal life generally, so still end up dominating total consciousness-space-time.)

The extra computing power available in the future might not mostly be used for individual integrated intelligent systems, but could be dispersed more widely, such as by being used to run simulations. So this would mean that while AI++s might end up existing, the small number of individual superintelligent systems could mean that lower intelligences such as us still take up a reasonable proportion of total consciousness-space-time.

It is a fairly popular idea that we ourselves could be part of a simulation (e.g. Bostrom, 2003), perhaps being run by a superintelligent AI. The idea is that once a computer is powerful enough, it can simulate a universe, including the life in it. And in this simulation there would potentially be further simulations and so on. If this is the case, then almost all universes would be simulated universes, meaning that we would almost certainly be living in a virtual universe ourselves. This could also mean a proliferation of human-level intelligence, which, along with the fact that these complex simulations would use computing power at the expense of AI++s, could counteract the potential consciousness-space-time dominance of AI++s.

However, a simulation can never fully represent a universe as complex as the one that it resides in, at least not in real time. To do so would require the same resources that the universe contains, but within a much smaller subset of that universe, which is not possible. The caveat about real time is important because a computer could simulate something more complex than itself, as long as we're not worried about it running much more slowly than real time. But we are. For any given amount of time, a computer running a simulation (or any number of simulations) cannot perform any more operations than it can when not running a simulation. There is no free lunch. This is also why computers emulated by your home PC are always far less powerful than the PC itself. A modern-day PC can emulate a computer from the 1980s, for example, but you'll never see it running an emulation of itself or a more powerful machine.

As a more extreme case, imagine if our universe was simulated on a computer inside our own universe. To be an accurate simulation, that universe would have to also contain a simulation of itself, and so on, leading to an infinite regress. It would therefore take an infinite amount of computing power to manage it. But to say that a computer cannot simulate the entire universe in real time is a vast understatement. It can't simulate the room it's sitting in in real time, because the room contains the computer being simulated, so this would still lead to the same infinite regress.

If we want to run a simulation of a universe, this can be a less complex universe than our own, or one that runs slowly. Either way, it means that for any finite amount of time we leave it running for, more will happen in "real life" than in the simulation, giving us reason to doubt that we are in a simulation ourselves.

Of course, this doesn't mean that we cannot be living in a simulated reality. But each simulation, and simulation of a simulation etc., is likely to have fewer and fewer conscious minds as the computing resources diminish the further along the line you go.

It is actually still possible to have a simulation that has more conscious beings than the reality it is represented in, by concentrating more of its power on the simulated beings themselves and less on the rest of the detail of the simulated universe. But it is likely that achieving this would give these beings a rather impoverished existence without much of a detailed outside world, and there is no reason to suggest that we are in such a simulation.

Even if some future computing power is dedicated to running simulations, given the likely limitations of such simulations, it is unlikely that any conscious beings in these simulations will prevent AI++s from dominating consciousness-space-time. By SSSSA, something is likely to prevent them, but we need to look beyond simulations.

I'm not going to try to consider every possibility of why we might create only a relatively small number of AI++s on the assumption that we develop the technology to be able to create them at all. It might be that there are other uses for future computing resources that sufficiently take away from AI++s. I discussed simulations specifically because it is an idea that has gained some currency. But the point is that even if one can envisage further scenarios to challenge premise 1, it still seems to be a fairly plausible premise at present. So the fact that SSSSA suggests that there probably won't be a proliferation of AI++s means that this should overall lower our credence that we will create a single AI++, unless we can find fault with premise 2.

## 6. Attacks on Premise 2 – Chinese Room

There is an argument that AI++s wouldn't actually be conscious, so their existence would have no effect on whether we are typical conscious observers or not. This is an attack on premise 2 above: that the creation of a large number of AI++s would cause them to dominate consciousness-space-time. John Searle's famous Chinese Room argument (e.g. 1980) is an attempt to demonstrate that no digital computer, however powerful, could be conscious.

Searle argues that everything that digital computers do is blind symbol manipulation, and that no amount of this can lead to real understanding, as goes on in our brains. Searle compares a computer to a person sitting inside a room who is given questions in Chinese, which are being passed through a hole. The person in the room understands no Chinese, but he has instructions, written in English, explaining what to do with the Chinese writing. Following the instructions leads him to produce the correct output, which is also in Chinese, although he still has no understanding of it. He passes the written output back out through the hole. So it is possible for someone outside the room to have a conversation with the Chinese Room, in Chinese, despite there being no-one in the room who understands Chinese. Searle says that there is no understanding of Chinese in the Chinese Room and, analogously, there is no understanding in a digital computer.

This argument is not generally accepted and I won't go into it in great detail here, although I discuss it in more detail in my book (Pereira, 2014). But in simple terms, individual neurons blindly fire without having any idea of the overall picture, and that does not seem to put a stop on human consciousness, and there is no obvious reason why blind symbol manipulation in a computer is any less likely to bring about consciousness than blind neuronal firings. The Systems Reply to the Chinese Room argument says that it's the system as a whole that is conscious rather than the individual units. And the person in the Chinese Room is just one such unit, just as the neurons are individual units in our brains. So the fact that the person in the Chinese Room has no conscious understanding of Chinese is irrelevant to whether there is any conscious understanding of Chinese at all, in the same way that the lack of consciousness in individual neurons in a normal human brain is irrelevant to whether there is any consciousness in this brain.

On top of this, Searle does accept that the human brain is a machine of sorts, but just not of the same type as a digital computer, so a conscious superintelligent AI could still be created by using the same or a similar method to our biology. So accepting Searle's Chinese Room argument does not mean that a conscious AI++ will not be created.

While potential challenges to premises 1 and 2 could come from many angles, their mere plausibility, along with acceptance of SSSSA, is enough to at least decrease our credence that we will create superintelligent AI.

## 7. Concluding Remarks

Nothing in this paper demonstrates that we will categorically not create superintelligent AI, but taking the central argument seriously should lead us to ask questions about the likely future of AI development. The argument itself also does not provide us with any particular obstacle to creating superintelligent AI; it merely suggests that there might be such an obstacle. It could be that we end up destroying ourselves, that it is much harder to create than some people imagine, or something else entirely. But to be clear, the implication is not merely that humans on Earth will probably not create superintelligent AI, but that when we consider human-level intelligent beings in the universe as a whole, and in any other universes that might exist, we should not expect enough of these races to develop superintelligent AI so that it dominates consciousness-space-time. The point is that we should expect human-like consciousness to take up a reasonable proportion of consciousness-space-time when considered as a whole, not just on Earth. Superintelligent AI may well be a rare development among intelligent life across the whole universe, and indeed beyond.